\documentclass[submission,copyright,creativecommons,sharealike]{eptcs}
\providecommand{\event}{LFMTP 2019}
\usepackage{breakurl}
\usepackage{underscore}

\usepackage{url,xspace,calbf,wrapfig,amstext,amssymb}
\usepackage{paralist}
\usepackage{hyperref}
\usepackage{MnSymbol}

\usepackage{courier}
\usepackage[utf8]{inputenc}
\usepackage{stmaryrd}  % contains replacement for MMT delimiter
\usepackage{caption}
\usepackage{tikz}
\usepackage{tikz-qtree}
\usetikzlibrary{mmt}

\usepackage{listings}

\lstdefinelanguage{GF}{
    morekeywords={ abstract, concrete, resource, interface, instance,
        incomplete, of, with, open, in,
        cat, fun, lincat, lin, oper, flags, param,
        def, lindef, linref, data
    },
    morecomment=[l]{--},
    morecomment=[s]{\{-}{-\}},
    morestring=[b]",
    showstringspaces=false,
    basicstyle=\footnotesize\ttfamily,
    captionpos=b
}

\def\mmtBasicStyle{\footnotesize\ttfamily}
\def\jOD{|}
\def\jDD{$\|$}
\def\jMD{$\talloblong$}
\def\jraa{$\rightarrow$}

\lstdefinelanguage{MMT}{
    morekeywords={ theory, view, include },
    basicstyle=\mmtBasicStyle,
    captionpos=b,
    literate={❘}{{\jOD}}1 {❙}{{\jDD}}1 {❚}{{\jMD}}1
             {⟶}{{\jraa}}1
             {⟪}{{$\llangle$}}1 {⟫}{{$\rrangle$}}1
             {⟦}{{$\llbracket$}}1 {⟧}{{$\rrbracket$}}1
             {∧}{{$\land$}}1 {¬}{{$\neg$}}1 {∨}{{$\lor$}}1
             {φ}{{$\varphi$}}1 {ψ}{{$\psi$}}1
             {ι}{{$\iota$}}1 {μ}{{$\mu$}}1
             {∪}{{$\cup$}}1 {∩}{{$\cap$}}1
             {∀}{{$\forall$}}1 {∃}{{$\exists$}}1
             {⊦}{{$\vdash$}}1
}

\lstdefinelanguage{BNR}{
    basicstyle=\footnotesize\ttfamily,
    stringstyle=\slshape,
    morestring=[b]",
    keepspaces=true
}

\def\gfinline#1{\lstinline[language=GF, breaklines=true, breakatwhitespace]{#1}}
\def\bnrinline#1{\lstinline[language=BNR, breaklines=true, breakatwhitespace]{#1}}
\def\mmtinline#1{\lstinline[language=MMT, breaklines=true, breakatwhitespace]{#1}}
\def\mmtstyle#1{{\mmtBasicStyle #1}}
\def\theorynode#1{\,\mmtstyle{#1}\,}

\def\GF{\textsf{GF}\xspace}
\def\MMT{\textsf{MMT}\xspace}
\def\GLF{\textsf{GLF}\xspace}
\def\GLFF#1{\textsf{GLF}(\ensuremath{\cF})}

\def\ommt{\textsf{OMDoc/MMT}\xspace}
\def\myemph#1{\textbf{#1}}
\def\str#1{``\textit{#1}''}   % Examples in natural language
\def\log#1{#1}  % logical statement example (not always used!)

\title{\GF + \MMT = \GLF \\ From Language to Semantics through LF}
\author{Michael Kohlhase \quad\quad Jan Frederik Schaefer \\ \institute{Computer Science, FAU Erlangen-N\"urnberg}}
\def\titlerunning{GF + MMT = GLF -- From Language to Semantics through LF}
\def\authorrunning{Michael Kohlhase \& Jan Frederik Schaefer}

\begin{document}
\maketitle
\begin{abstract}
    These days, vast amounts of knowledge are available online,
    most of it in written form.
    Search engines help us access this knowledge,
    but aggregating, relating and reasoning with it is still a predominantly human effort.
    One of the key challenges for automated reasoning based on natural-language texts
    is the need to extract meaning (semantics) from texts.
    Natural language understanding (NLU) systems
    describe the conversion from a set of natural language utterances to terms in a particular logic.
    Tools for the co-development of grammar and target logic are currently largely missing.

    We will describe the \myemph{Grammatical Logical Framework} (\GLF),
    a combination of two existing frameworks, in which large parts of a symbolic, rule-based NLU system can be developed and  implemented:
    the Grammatical Framework (\GF) and \MMT.
    \GF is a tool for syntactic analysis, generation, and translation with complex natural language grammars
    and \MMT can be used to specify logical systems and to represent knowledge in them.
    Combining these tools is possible, because they are based on compatible logical frameworks: Martin-L\"of type theory and LF.
    The flexibility of logical frameworks is needed, as NLU research has not settled on
    a particular target logic for meaning representation. Instead, new logics are developed all the time to handle various language phenomena.
    \GLF allows users to develop the logic and the language parsing components in parallel, and to
    connect them for experimentation with the entire pipeline.
\end{abstract}

\section{Introduction}

\myemph{Natural language semantics} studies the meaning of natural language utterances.  A
fundamental conceptual tool for this are \myemph{truth conditions}: the set of conditions
under which an NL utterance is true.  For example, \str{John loves Mary} is true if and only
if John indeed loves Mary.  Somewhat less tautologously: two assertions have the same meaning, if they have the same truth
conditions.  We can therefore identify the meaning of an assertion with its truth
conditions~\cite{Davidson:tam67}. This notion of ``meaning'' is very general, but also not
very constructive. Therefore truth conditions are generally thought of as a minimal
axiomatization of the domains of discourse which entails the assertion.

\providecommand\NLlang{\mathcal{NL}}
\providecommand\LogicFL{\mathcal{FL}}
\providecommand\NLentailmentRel{\models_{\mathcal{NL}}}
\begin{wrapfigure}r{8.9cm}\centering\vspace*{-1.5em}
  \begin{tikzpicture}[yscale=.7]
    \draw[line width=1.5pt,rounded corners=.3cm,fill=green!25] (-1.7,-1) rectangle (1.5,3.5);
    \node (cl) at (0,-.5) {Comp Ling};
    \node (nl) at (0,0) {$\NLlang$};
    \node (L) at (0,2) {$\LogicFL$};
    \node (M) at (0,4) {$\cM=\langle\cD,\cI\rangle$};
    \node (Inf) at (6,0) {$\models_\NLlang \; \subseteq \NLlang \times \NLlang$};
    \node (C) at (6,2) {$\vdash_\cC \; \subseteq \LogicFL\times\LogicFL$};
    \node (folg) at (6,4) {$\models \; \subseteq \LogicFL\times\LogicFL$};
    \draw[->] (nl) -- node[left] {Analysis} (L);
    \draw[->] (L) -- node[left] {$\mathcal{I}_\phi$} (M);
    \draw[dotted,->] (nl) -- node[above] {induces} (Inf);
    \draw[dotted,->] (M) -- node[above] {induces} (folg);
    \draw[->] (L) -- node[above] {formulae} (C);
    \draw[<->] (folg) -- node[left]{$\models \; \equiv \; \vdash_\cC$?} (C);
    \draw[<->] (C) -- node[left] {$\models_\NLlang \; \equiv \; \vdash_\cC$?} (Inf);
  \end{tikzpicture}
  \caption{ Natural-language inference on different levels.}\label{fig:NLinference}\vspace*{-1em}
\end{wrapfigure}
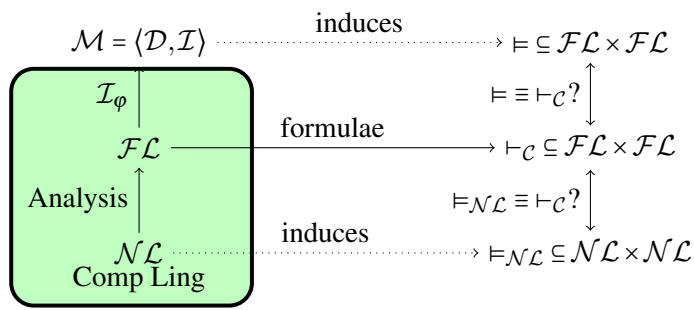

To understand this setup, assume that we use a formal language $\LogicFL$ to express truth conditions.
The meaning of \str{John loves Mary} could then be \log{$\textit{love}(\textit{john}, \textit{mary})\in\LogicFL$}.
If $\cF\cL$ is the formal language of a logical system, we also have an interpretation function $\cI_\varphi$ of $\cF\cL$ expressions into a model and a calculus $\cC$ with a derivation relation $\vdash_\cC$.
If $\cC$ is sound and complete, the upper rectangle in Figure~\ref{fig:NLinference} commutes.
If the calculus $\cC$ is an adequate model for the natural language entailment relation $\models_{\NLlang}$ -- also called ``textual entailment'' in the linguistics literature -- both rectangles in Figure~\ref{fig:NLinference} commute and we have a good model for truth conditions and logical entailment for natural language utterances.
In this case, it suffices to specify the translation from natural language $\NLlang$ to the formal language $\LogicFL$ along with a calculus $\cC$.
And in general, the ``NL semantics'' literature restricts itself to the box in Figure~\ref{fig:NLinference}, entrusting the upper square to logicians and the equivalence of $\cM$-entailment and textual entailment $\models_\NLlang$ to the logic developers.
At the same time, NL semanticists continually need extensions to $\LogicFL$ and $\cC$ to model new NL phenomena.

In particular, it is still unachievable to describe a translation from the entirety of natural language into some formal language.
Instead, researchers rather focus on particular phenomena in natural language by describing a small subset of natural language utterances (a \myemph{fragment}) along with the meaning of these utterances.
This \myemph{method of fragments} was established by Richard Montague~\cite{Montague:efl70}.
It typically results in the description of three components:
\begin{compactenum}
\item a \myemph{grammar} that fixes the language fragment and generates syntax trees
\item a \myemph{formal system} in which the semantics of utterances can be expressed
\item a way to transform syntax trees to expressions in the formal system,
  which is often referred to as \myemph{semantics construction}
\end{compactenum}

The semantics construction is based on the \myemph{compositionality principle}:
the idea that the meaning of a complex utterance is determined by the meaning of its constituents.
Thereby, the semantics construction boils down to mapping grammar rules to corresponding semantic operations.
Consider, for example, the grammar rule \bnrinline{<sentence> ::= <sentence> "and" <sentence>}.
It corresponds to the semantic operation \log{$\llbracket A \rrbracket \land \llbracket B \rrbracket$}, where \log{$\llbracket A \rrbracket$} and \log{$\llbracket B \rrbracket$} are the meanings of the constituent sentences.
The semantics construction may be followed by a \myemph{semantic analysis}\footnote{
    In Anglo-Saxon literature this is sometimes called \myemph{pragmatics}.
}, which comprises various non-compositional operations such as inference, anaphora resolution, or contextual anchoring.

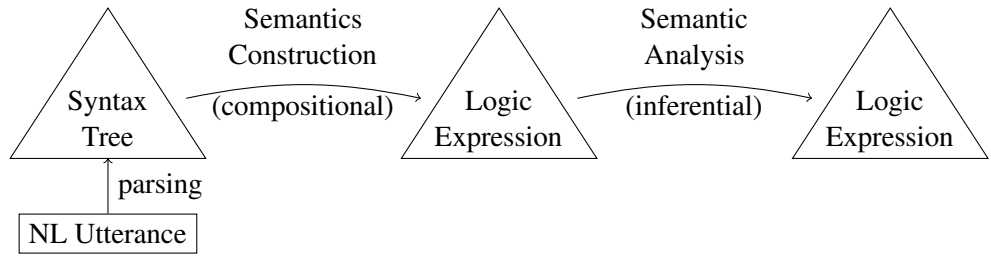
\begin{figure}\centering
    \begin{tikzpicture}[xscale=1.3]
      % \node at (0,3.5) {\textbf{\large Syntax}};
      % \node at (4,3.5) {\textbf{\large Quasi-Logical Form}};
      % \node at (8,3.5) {\textbf{\large Logical Form}};
      \node[draw,minimum width=2cm] (utt) at (0,0) {NL Utterance};
      \draw[] (-1,1) -- (0,3) -- (1,1) -- cycle;
      \node[] (st) at (0,1.5) {\begin{tabular}{c}Syntax\\Tree\end{tabular}};
      \draw[->] (utt) -- node[right] {parsing} (0,1);
      \draw (3,1) -- (4,3) -- (5,1) -- cycle;
      \node (qlf) at (4,1.5) {\begin{tabular}{c}Logic\\Expression\end{tabular}};
      \draw[->] (0.8,1.8) to[bend left=15] node[above] {\begin{tabular}{c}Semantics\\ Construction\end{tabular}} node[below]{(compositional)} (3.2,1.8);
      \draw (7,1) -- (8,3) -- (9,1) -- cycle;
      \node (le) at (8,1.5) {\begin{tabular}{c}Logic\\Expression\end{tabular}};
      \draw[->] (4.8,1.8) to[bend left=15] node[above] {\begin{tabular}{c}Semantic\\ Analysis\end{tabular}} node[below]{(inferential)} (7.2,1.8);
    \end{tikzpicture}
    \caption{The pipeline of an NLU system.}\label{fig:NLUpipeline}
\end{figure}

Symbolic \myemph{natural-language understanding (\emph{NLU}) systems} describe the entire pipeline from strings to semantic representations (Figure~\ref{fig:NLUpipeline}).
They have been used to describe the semantics of a variety of natural-language phenomena.
In the process, many different logics have been developed.
However, the experiments were mostly done with pen and paper, and have rarely been implemented in software.
This can lead to researchers focusing either on the linguistic side or on the logical side of the problem, while the actual semantics construction remains vague.

When someone actually implemented such an NLU system, it was usually done in a programming language like Prolog or Haskell -- see e.g.~\cite{BlaBos:rainl05,EijUng:csfp10}.
In both cases, the authors claim that the programming language is an NLU framework -- in the first case since Prolog is a declarative programming language and in the second since Haskell is very high-level.
In any case, the NLU system  requires a considerable -- potentially prohibitive -- amount of programming work.
As far as we can tell, there is no fully declarative framework that could be used to do both, the grammar development as well as the logic development and keep them in sync.
% \footnote{The closest to such a framework
% is probably the Grammatical Framework described in Section~\ref{sec:gf}.
% However, we claim that more powerful logic development possibilities are desirable.}

In this paper we describe our efforts to create the \myemph{Grammatical Logical Framework} (\GLF).
It combines an existing framework for natural language grammars with an existing
framework for logic development.
Concretely, we combine the \myemph{Grammatical Framework} (\GF)~\cite{ranta-2011},
with the logic development tool \MMT~\cite{RabKoh:WSMSML13}.
This is possible, because the logical frameworks underlying these tools are compatible.
\GF handles the natural language parsing and generates terms (parse trees) in a logical
framework (Martin-L\"of type theory~\cite{Ranta:GF04}).
\MMT, which supports LF and various extensions,
maps these terms to expressions in the desired target logic -- see Figure~\ref{fig:glff}.

In GLF, an NLU researcher can specify a fragment of a language in \GF and, in parallel, develop a logic in \MMT, along with a domain theory and the semantics construction.
Our framework supports this in various ways, such as:
\begin{compactitem}
\item it allows the researcher to try out the entire pipeline from an utterance to its logical representation
\item it checks the totality of the semantics construction
\item the grammar and the logic are type-checked as usual in \GF and \MMT
\end{compactitem}
We admit that symbolic natural language understanding has dropped from the limelight of computational linguistic research in the last two decades in favor of machine-learning-based approaches.
But the success of these  has only shadowed the question of semantic analysis and natural language inference.
We see a cautious revival of symbolic/logic-based methods in computational linguistics, and we hope that \GLF can serve as a tool to facilitate this.  

The symbolic approach to NLU needs extensive resources (e.g. grammars and ontologies).
Aarne Ranta, the creator of \GF, distinguish two areas of NL applications:
\myemph{consumer tasks} and \myemph{producer tasks}~\cite{Ranta:GfGoogle2016}.
Consumer tasks require large coverage -- often achieved through machine learning --
and are therefore typically limited in their precision.
An example of this would be machine learning-based text translation a la Google Translate.
Producer tasks, on the other hand, require high precision, but are restricted in their coverage
to a few thousand concepts.
An example are technical manuals for complex machinery in dozens of languages,
where the consequences of mistranslation may be catastrophic.
Beyond translation, producer tasks -- the natural hunting grounds of \GLF --
include understanding of mathematical papers, laws or contracts.

\paragraph{Overview}
First, we will describe \GF and \MMT (Sections~\ref{sec:gf} and~\ref{sec:mmt}).
After an overview of the \GLF system (Section~\ref{sec:glfSystem}), we will describe the semantics construction and semantic analysis (Sections~\ref{sec:semConstr} and~\ref{sec:semAnal}) using a running example.
Section~\ref{sec:examples} contains more examples of how this framework can be used.
Section~\ref{sec:concl} concludes the paper and discusses future work. 

\paragraph{Acknowledgements}
We are grateful for the discussions with and insights from Aarne Ranta, Florian Rabe, and finally Dennis M\"uller, who has also prototyped an early version of \GLF.
The work reported here was supported by the German Research Foundation (DFG) under grant KO 2428/18. 

\section{\GF: The Grammatical Framework}\label{sec:gf}
The \myemph{Grammatical Framework} (\GF)~\cite{ranta-2011, GF:on}
can be used to create multilingual grammar applications.
\GF grammars are divided into two parts: \myemph{abstract syntax}
and \myemph{concrete syntaxes}.
The abstract syntax describes the \myemph{ASTs} (\myemph{abstract syntax trees} or
\myemph{abstract syntax terms}) covered by the grammar.
The concrete syntaxes are AST linearization rules in a specific natural language.

Let us  consider a small example:
Listing~\ref{lst:GF-Life} shows an abstract syntax for representing some statements about everyday life such as \str{Joan runs and Mary loves Joan}.
First, three basic types are introduced (\gfinline{Stmt}, \gfinline{Person}, \gfinline{Action}) with the keyword \gfinline{cat} (in \GF they are called \myemph{categories}).
Afterwards, several function constants are introduced with the keyword \gfinline{fun}.
The example utterance \str{Mary loves Joan} would correspond to the AST \gfinline{act mary (love joan)}.
Formally, \GF is based on a version of constructive type theory~\cite{Ranta:GF04}.
It supports dependent types, but, in our experience, these are not very useful for most natural-language grammar applications.

\begin{figure}[ht]\centering
\begin{minipage}[t]{.5\textwidth}
\begin{lstlisting}[language=GF]
abstract Life = {
  cat
    Stmt ; Person ; Action ;
  fun
    act : Person -> Action -> Stmt ;
    and : Stmt -> Stmt -> Stmt ;

    joan        : Person ;
    mary        : Person ;
    run         : Action ;
    loveOneself : Action ;
    love        : Person -> Action ;
}
\end{lstlisting}
\end{minipage}\hfill
\begin{minipage}[t]{.5\textwidth}
\begin{lstlisting}[language=GF]
concrete LifeEng of Life = {
  lincat
    Stmt, Person, Action = Str ;
  lin
    act pers action = pers ++ action ;
    and a b         = a ++ "and" ++ b ;

    joan        = "Joan" ;
    mary        = "Mary" ;
    run         = "runs" ;
    loveOneself = "loves" ++ "herself" ;
    love person = "loves" ++ person ;
}
\end{lstlisting}
\end{minipage}\vspace*{-1em}
\captionof{lstlisting}{Abstract syntax to talk about ever-day life along with English linearization rules.}\label{lst:GF-Life}
\end{figure}

GF's concrete syntaxes describe how the ASTs are linearized in a particular natural language.
Listing~\ref{lst:GF-Life} shows a concrete syntax for the English language.
First, the linearization types of the categories are defined (\gfinline{lincat}).
\GF provides a powerful type system for the concrete syntax with record types and more.
For this example the basic type \gfinline{Str} suffices;
this would change if a male person is added,
since we would have to distinguish between \str{loves herself} and \str{loves himself}.
The keyword \gfinline{lin} is used to describe the linearizations of function constants.

With the concrete syntax, ASTs can be linearized into English strings.
\GF also generates a parser from the linearization rules, which allows us to parse English strings into ASTs.
By creating another concrete syntax, e.g. for German, we can parse strings in one language and linearize the ASTs into a different language.
This allows us to translate between languages.
For instance the invocation 
\begin{lstlisting}[language=GF,keywords={parse,linearize},stringstyle=\itshape]
parse -lang=Eng -cat=Stmt "Mary loves herself" | linearize -lang=Ger
\end{lstlisting}
in the \GF shell yields the result
\begin{lstlisting}[language=GF]
Maria liebt sich
\end{lstlisting}

The example above is of course a very simple one and does not contain any of the challenges
one would typically face in a natural-language grammar.
Any more complex natural-language application will
have to implement a large number of rules for handling the word order and word forms.
In English, for example, verb endings depend on the plurality of the subject.
These rules can be implemented in \GF's concrete syntaxes using records, tables, etc.
Since these language-specific rules are needed in most projects,
the \GF community provides the \myemph{Resource Grammar Library}~\cite{RGL:on}.
It contains rules for currently 36 languages
and is an invaluable resource for creating natural-language applications.

\section{\MMT: Meta-Meta Theories/Tool}\label{sec:mmt}
\ommt (Meta-Meta Theories) is a modular, foundation-independent framework for representing knowledge~\cite{RabKoh:WSMSML13,MueRab:rpfsm}.
The \MMT system (Meta-Meta Tool~\cite{Rabe:MAGMS13,uniformal:on}) acts as an \ommt API and implements various knowledge management services including type/proof checking.  

\ommt knowledge is represented in \myemph{theories},
which contain (among other things) declarations of constants.
A constant declaration \mmtstyle{c[:$\tau$][=$\delta$][\#$\nu$]}
can have various components such as a type $\tau$, a definiens $\delta$ or a notation $\nu$.

\ommt is a modular framework: domain ontologies and logics are represented as graphs of theory presentations inter-linked by \textbf{theory morphisms} that model inheritance and interpretation.
In practice \ommt development follows a ``little theories'' paradigm, where each theory presentation only contributes a small number of declarations to maximize re-use of content. 

Due to the foundation-independence, various logical frameworks can be implemented in \MMT, but in practice LF and extensions are usually sufficient to represent a wide range of formal systems in \MMT.  See~\cite{CodHorKoh:palai11,LATIN:git} for the LATIN Logic Atlas and Integrator, a large modular theory graph of LF-encoded logics. 

As an example, let us create an \MMT theory \mmtinline{PropLogicSyntax} for propositional logic, which we can base on LF.
Note that LF is just another theory in \MMT, the \myemph{meta theory} of \mmtinline{PropLogicSyntax}.
The theory \lstinline{LF} provides $\lambda$ for functions, $\Pi$ for dependent types, the kind \mmtinline{type} , and the function type constructor $\rightarrow$.
Apart from a meta theory, \mmtinline{PropLogicSyntax} has four constant declarations (Listing~\ref{lst:MMT-PropLogSyn}).
First, the constant \mmtinline{prop} is declared as a type with the notation \mmtinline{o} (the symbol \mmtinline{#} introduces the notation).
The components of a constant declaration are separated by the delimiter \mmtstyle{\jOD}, declarations with \mmtstyle{\jDD}, and theories/views with \mmtstyle\jMD.
We can declare conjunction (\mmtinline{and}) and negation (\mmtinline{neg}) as binary/unary operations on propositions using the notation \mmtinline{o} introduced for \mmtinline{prop}.
Applying De Morgan's law, disjunction (\mmtinline{or}) can be defined in terms of conjunction and negation.
Note that in \MMT lambda expressions like $\lambda x, y.M$ are written as \mmtinline{[x,y] M}.

\begin{figure}[h]
    \begin{minipage}[t]{0.5\textwidth}
\begin{lstlisting}[language=MMT]
theory PropLogicSyntax : ur:?LF =
    prop : type      ❘ # o ❙
    and  : o ⟶ o ⟶ o ❘ # 1 ∧ 2 ❙
    neg  : o ⟶ o     ❘ # ¬ 1 ❙
    or   : o ⟶ o ⟶ o ❘ # 1 ∨ 2 ❘
         = [x,y] ¬ ((¬ x) ∧ (¬ y)) ❙
❚
\end{lstlisting}
    \end{minipage}\hfill
    \begin{minipage}[t]{0.5\textwidth}
\begin{lstlisting}[language=MMT]
theory PropLogicND : ur:?LF =
    include ?PropLogicSyntax ❙
    ded : o ⟶ type ❘ # ⊦ 1 ❙

    andI  : {A,B} ⊦ A ⟶ ⊦ B ⟶ ⊦ (A ∧ B) ❙
    andEr : {A,B} ⊦ (A ∧ B) ⟶ ⊦ B ❙
    ...
\end{lstlisting}
    \end{minipage} % label={lst:MMT-PropLogSyn}
    \captionof{lstlisting}{Propositional logic in \MMT: syntax (left) and proof theory (right).}\label{lst:MMT-PropLogSyn}
\end{figure}
The right side of Listing~\ref{lst:MMT-PropLogSyn} shows the usual representation of the natural deduction calculus for propositional logic via the Curry-Howard isomorphism.
We will use it for the verification of truth conditions in domain theories below.

To model the semantics of propositional logic in set theory, we create a new theory \mmtinline{PropLogicModel} (Listing~\ref{lst:proplogsem}) that is based on set theory (\mmtinline{sets:?AllSets}).
In \ommt terms, set theory is the meta theory of \mmtinline{PropLogicModel}.
The meta theory relation is a theory morphism, very similar to inclusion/inheritance.

We want to interpret propositions as sets of satisfying variable assignments and interpret falsity as the empty set, therefore
\mmtinline{PropLogicModel} has a single constant declaration, which introduces a new type constant with the notation \mmtinline{A} for variable assignments.

\begin{figure}[h]
    \begin{minipage}[t]{0.45\textwidth}
\begin{lstlisting}[language=MMT]
theory PropLogicModel :
            sets:?AllSets =
    assignment : type ❘ # A ❙
❚
\end{lstlisting}
    \end{minipage}\hfill
    \begin{minipage}[t]{0.65\textwidth}
\begin{lstlisting}[language=MMT]
view PropLogicSemantics :
        ?PropLogicSyntax -> ?PropLogicModel =
    prop = set A ❙
    and = [φ, ψ] φ ∩ ψ ❙
    neg = [φ] (fullset A) \ φ ❙
❚
\end{lstlisting}
    \end{minipage}
    \captionof{lstlisting}{Semantics description as view (right) into \mmtinline{PropLogicModel} (left).}\label{lst:proplogsem}
\end{figure}

\begin{figure}[ht]\centering
    \begin{tikzpicture}[xscale=1.5,yscale=1]
        \node[thy] (lf) at (-2.5,2.5)  {\theorynode{ur:?LF}};
        \node[thy] (set) at (2.5,2.5)  {\theorynode{sets:?AllSets}};

        \node[thy] (pls) at (-2.5,1.0) {\theorynode{PropLogicSyntax}};
        \node[thy] (plm) at (2.5,1.0) {\theorynode{PropLogicModel}};

        \draw[meta] (lf) -- (pls);
        \draw[meta] (set) -- (plm);

        \draw[view] (pls) -- node[above]{\theorynode{PropLogicSemantics}} (plm);
    \end{tikzpicture}
    \caption{Propositional Logic as an \ommt Theory Graph.}\label{fig:prop-log}
\end{figure}
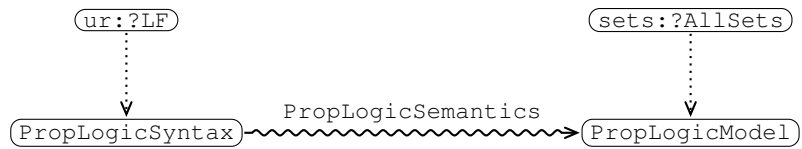

The meaning of \mmtinline{PropLogicSyntax} can now be established with another kind of theory morphism: a \myemph{view} maps undefined constants from the source theory to objects in the target theory.
In this case, we will describe the meaning of \mmtinline{PropLogicSyntax} with a view
from \mmtinline{PropLogicSyntax} to \mmtinline{PropLogicModel} (Listing~\ref{lst:proplogsem}).
A proposition can be represented as the set of variable assignments that
make it true.
Then, the conjunction of two propositions corresponds to the intersection of two sets
and negation corresponds to taking the complement set.
As disjunction (\mmtinline{or}) is defined in terms of conjunction and negation,
we do not need to map it to anything here.
Figure~\ref{fig:prop-log} provides an overview of the different theories and morphisms
we have used.

\section{\GLF: The Grammatical Logical Framework}\label{sec:glfSystem}
\GLF combines \GF and \MMT by exploiting the compatibility of the underlying logical frameworks.
In the next sections we will explain how this leads to a framework for natural-language semantics.
We start with the information and software architecture of the system.

\subsection{\GF vs \MMT{} -- Two Sister Formats}\label{sec:sisterformat}
% % \GF will serve as a grammar development tool and \MMT as a logic/semantics development tool.
% % \GF's parse trees are terms in a logical framework which are compatible with \MMT,
% % which allows us to write an \MMT theory equivalent to the abstract grammar in \GF.
% % Converting a \GF parse tree to an \MMT object is then trivial.
% % For the semantics, we can use \MMT to develop a logic and create a domain theory.
% % The semantics construction is then achieved with \MMT's views
% % by mapping parsed terms into the logic and domain theory.

% As mentioned before, our goal is to create a NL semantics development framework, \GLF, by combining the Grammatical Framework (\GF) and \MMT.
% For this, we will exploit that \GF's abstract syntaxes
% can be trivially represented as \MMT/LF theories, due to the compatibility of the underlying logical frameworks.
% We will refer to an \MMT theory that corresponds to an abstract syntax as a \myemph{language theory}.
\GLF's combination of \GF and \MMT is based on the fact that \GF's abstract syntaxes
can be trivially represented as \MMT/LF theories.
We will refer to an \MMT theory that corresponds to a \GF abstract syntax module as a \myemph{language theory}.

We can illustrate this using the \gfinline{Life} grammar from Listing~\ref{lst:GF-Life}.
For instructive reasons, we will split it into two parts:
a \gfinline{LifeGrammar} that deals with the ``grammatical'' structures of a language and a \gfinline{LifeLex} that contains the lexical entries.
Listing~\ref{lst:LifeLex} shows the \gfinline{LifeGrammar} in \GF, along with its language theory in \MMT.
\GF categories correspond to type constants and \GF functions to function constants in \MMT.
Listing~\ref{lst:LifeLex} also shows the \gfinline{LifeLex} abstract syntax in \GF with the corresponding language theory.
Note that \gfinline{LifeLex} extends \gfinline{LifeGrammar} in \GF, and correspondingly in \MMT \mmtinline{LifeLex} includes \mmtinline{LifeGrammar}.
Generally, the \MMT module system subsumes the one of \GF, thus we can build the information architecture in parallel.  

\begin{figure}
\begin{minipage}[t]{.5\textwidth}
\begin{lstlisting}[language=GF]
abstract LifeGrammar = {
  cat
    Stmt ;
    Person ;
    Action ;
  fun
    act : Person -> Action -> Stmt ;
    and : Stmt -> Stmt -> Stmt ;
}
\end{lstlisting}
\end{minipage}\hfill
\begin{minipage}[t]{.5\textwidth}
\begin{lstlisting}[language=MMT]
theory LifeGrammar : ur:?LF =

    Stmt   : type ❙
    Person : type ❙
    Action : type ❙

    act : Person ⟶ Action ⟶ Stmt ❙
    and : Stmt ⟶ Stmt ⟶ Stmt ❙
❚
\end{lstlisting}
\end{minipage}

\begin{minipage}[t]{.5\textwidth}
\begin{lstlisting}[language=GF]
abstract LifeLex = LifeGrammar ** {
  fun
    joan        : Person ;
    mary        : Person ;
    run         : Action ;
    loveOneself : Action ;
    love        : Person -> Action ;
}
\end{lstlisting}
\end{minipage}\hfill
\begin{minipage}[t]{.5\textwidth}
\begin{lstlisting}[language=MMT]
theory LifeLex : ur:?LF =
    include ?LifeGrammar ❙
    joan        : Person ❙
    mary        : Person ❙
    run         : Action ❙
    loveOneself : Action ❙
    love        : Person ⟶ Action ❙
❚
\end{lstlisting}
\end{minipage}
    \captionof{lstlisting}{The \gfinline{LifeGrammar} and its extension \gfinline{LifeLex} in \GF (left) and \MMT (right).}\label{lst:LifeLex}    
\end{figure}

The similarity between \GF abstract syntaxes and their language theories in \MMT enables us to leave \GF after parsing and continue with the semantics construction in \MMT (Section~\ref{sec:semConstr}).
Specifically, an AST in \GF like \gfinline{act joan (love mary)} can trivially be mapped to the \MMT term \mmtinline{act joan (love mary)}.

\subsection{The \GLF System}\label{sec:system}
\GLF is a relatively thin wrapper around the unchanged \GF and \MMT systems: \GF is started in server mode and \GLF communicates with it via HTTP requests -- this turned out to be easier to set up than using the \GF Java bindings. The bridge between \GF and \MMT consists of

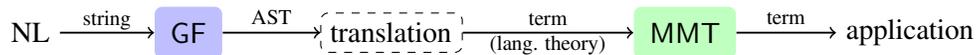
\begin{figure}[ht]\centering
  \begin{tikzpicture}[xscale=3.0]
    \node (nl) at (0.4,0) {NL};
    \node[rounded corners=.1cm,fill=blue!25,inner sep=6pt] (gf) at (1.1,0) {\GF};
    \node[draw,rounded corners=.1cm,dashed] (s) at (2,0) {translation};
    \node[rounded corners=.1cm,fill=green!25,inner sep=6pt] (m) at (3.3,0) {\MMT};
    \node (a) at (4.3,0) {application};
    \draw[->,line width=0.7pt] (nl) -- node[above=-3pt]{\scriptsize string} (gf);
    \draw[->,line width=0.7pt] (gf) -- node[above=-1pt]{\scriptsize AST} (s);
    \draw[->,line width=0.7pt] (s) -- node[anchor=center]{\scriptsize\begin{tabular}{c}term\\{(lang. theory)}\end{tabular}} (m);
    \draw[->,line width=0.7pt] (m) -- node[above=-1pt]{\scriptsize term} (a);
  \end{tikzpicture}
  \caption{The general \GLF pipeline.}\label{fig:glf}
\end{figure}

\begin{compactenum}
\item a small script that given an abstract \GF grammar $\cG$ generates the language theory $\widehat\cG$ and a stub for the semantics construction view $\widetilde\cG$ (cf. Section~\ref{sec:semConstr}) and
\item a translator that translates $\cG$-ASTs in \GF into $\widehat\cG$-terms in \MMT (cf. Section~\ref{sec:semConstr}).
\end{compactenum}
Figure~\ref{fig:glf} shows the pipeline: first, an utterance is parsed by \GF into an AST, which \GLF translates into an equivalent \ommt term, which \MMT can use for the semantics construction and analysis and pass on to an application. 
We can see this as a refinement/implementation of Figure~\ref{fig:NLUpipeline}.

\subsection{\GLF Fragments}\label{sec:GLF-frags}
Together, the \GF grammars and \MMT theories constitute a full formal basis for parsing and inference, therefore we call it a \textbf{\GLF Fragment}. 
Even though this example is very basic and didactically motivated, the general structure of a \GLF fragment is the same. It consists of:
\begin{compactenum}
\item an abstract \GF/MMT syntax, such as the one in Listing~\ref{lst:LifeLex},
\item a set of concrete \GF syntaxes for them, as in Listing~\ref{lst:GF-Life} on the right,
\item an \MMT logic, such as the one in Listings~\ref{lst:MMT-PropLogSyn} (syntax) and ~\ref{lst:proplogsem} (semantics), and
\item an \MMT view for the semantics construction, such as the one in Listing~\ref{lst:MMT-LifeSemantics-Split}.
\end{compactenum}

We write \GLFF\cF{} for \GLF with fragment $\cF$ loaded.  As we can directly experiment with a \GLF package $\cF$, \GLFF\cF{} can be used as a fragment development system.

Figure~\ref{fig:glff} refines the pipeline from Figure~\ref{fig:glf} taking the components of the \GLF fragment $\cF$ into account:
an utterance is parsed by \GF with one of the concrete syntaxes to obtain an AST (specified by the abstract syntax). Different concrete syntaxes correspond to different languages.
Afterwards, \GLF translates the AST into an equivalent \MMT term in the language theory.
Semantics construction, via \MMT (cf. Section~\ref{sec:semConstr}), and semantic analysis in \MMT (Section~\ref{sec:semAnal}) follow.

\begin{figure}\centering
    \begin{tikzpicture}[xscale=1.1]
    \draw[white,line width=0,rounded corners=.2cm,fill=blue!25] (-4,0.25) rectangle (0,3.75);
    \draw[white,line width=0,rounded corners=.2cm,fill=green!25] (0.5,0.25) rectangle (5,3.75);
%    \draw[gray,line width=1.5pt,rounded corners=.2cm,fill=blue!25] (-4,0.25) rectangle (0,4.25);
%    \draw[gray,line width=1.5pt,rounded corners=.2cm,fill=green!25] (0.5,0.25) rectangle (6,4.25);
    \node (nl1) at (-3,-0.5) {Lang. 1};
    \node (nli) at (-2,-0.5) {$\cdots$};
    \node (nln) at (-1,-0.5) {Lang. $n$};
    \node[draw,line width=0.7pt,inner sep=-0.0pt,rounded corners=2pt] (cs1) at (-3,1) {\small \begin{tabular}{c}Concrete\\Syntax 1\end{tabular}};
    \node (csi) at (-2,1) {$\cdots$};
    \node[draw,line width=0.7pt,inner sep=-0.0pt,rounded corners=2pt] (csn) at (-1,1) {\small \begin{tabular}{c}Concrete\\Syntax $n$\end{tabular}};
    \node[draw,line width=0.7pt,inner sep=-0.0pt,rounded corners=2pt] (as)  at (-2, 3.0) {\small \begin{tabular}{c}Abstract\\Syntax\end{tabular}};
    \draw[->,line width=0.7pt] (nl1) -- (cs1);
    \draw[->,line width=0.7pt] (nln) -- (csn);
    \draw[->,line width=0.7pt] (cs1) -- (as);
    \draw[->,line width=0.7pt] (csn) -- (as);

    \node[draw,line width=0.7pt,inner sep=-0.0pt,dashed,rounded corners=2pt] (lt) at (2, 3.0) {\small \begin{tabular}{c}Language\\Theory\end{tabular}};
    \node[draw,line width=0.7pt,inner sep=-0.0pt,rounded corners=2pt] (tl) at (4, 3.0) {\small \begin{tabular}{c}Target\\Logic\end{tabular}};
    \node[draw,line width=0.7pt,inner sep=-0.0pt,rounded corners=2pt] (dt) at (3, 1) {\small \begin{tabular}{c}Domain\\Theory\end{tabular}};

    \draw[view] (lt) -- node[sloped,anchor=center]{\footnotesize \begin{tabular}{c}Sem.\\Constr.\end{tabular}} (dt);
        \draw[view] (dt.180) arc (15:345:15pt) node[pos=0.5,right=-4pt]{\footnotesize \begin{tabular}{c}Sem.\\Anal.\end{tabular}};
    \draw[meta] (tl) -- (dt);

    \node[draw,line width=0.7pt,inner sep=-0.0pt,rounded corners=2pt] (at) at (6.5,1) {\small \begin{tabular}{c}Application\\Theory\end{tabular}};
        \draw[->,line width=0.7pt] (as) -- node[above]{\footnotesize generates} (lt);
    \draw[view,line width=0.7pt] (dt) -- node[sloped,anchor=center]{\footnotesize \begin{tabular}{c}Application\\Interpretation\end{tabular}} (at);

    \node (gf) at (-2,4.25) {\textbf\GF};
    \node (mmt) at (3,4.25) {\textbf\MMT};
  \end{tikzpicture}
    \caption{The \GLFF\cF{} Pipeline.}\label{fig:glff}
\end{figure}

% \GLF usually needs to be linked to an \MMT installation.

\subsection{\GLF Applications}\label{sec:GLF-appl}
We provide a small \GLF distribution at~\cite{GLFdemo} that
comes with \MMT and the grammars and semantics constructions described in this paper.
Concretely, this allows the reader to try out the examples in the command-line
by entering sentences and receiving the result of the semantics construction.
Here is an abbreviated example interaction based on an example extension of the \gfinline{Life} fragment
described in Section~\ref{sec:fol}:

\begin{lstlisting}[language=MMT]
Please enter a sentence: John and Mary run
I got the following interpretations:
(run' john')∧(run' mary')
Please enter a sentence: John loves everyone
I got the following interpretations:
∀[x:ι]love' john' x
\end{lstlisting}

Note that the example above doesn't require any implementation work at all -- only a \GLF fragment,
which gets passed to \GLF via command-line arguments.

In Section~\ref{sec:semAnal} we will motivate the use of tableaux for semantic analysis.
There is also a small demo for that, which allows the user to enter sentences and outputs the updated belief state.
Here is an abbreviated example interaction:

\begin{lstlisting}[language=MMT]
Please enter a sentence: if John doesn't love Mary then Mary doesn't love John
Here is my belief state:
  Option 1: love' john' mary'
  Option 2: NOT love' mary' john'
Please enter a sentence: John doesn't love Mary
Here is my belief state:
  Option 1: NOT love' mary' john' ;  NOT love' john' mary'
\end{lstlisting}

\section{Semantics Construction}\label{sec:semConstr}

The key observation of this paper is that semantics construction in Montagovian settings can be formalized in \ommt as a view from the language theory of a fragment $\cF$ to the corresponding domain theory.
\GLF generates the language theory in $\cF$ from the abstract grammar in $\cF$ and translates ASTs to $\MMT$, so that we can execute the semantics construction step by $\MMT$ functionality: view application and simplification.

Let us turn back to our example to fortify our intuition.
We need to define a suitable target logic and a domain theory.
For this small example we don't need a powerful logic:
it suffices to extend the propositional logic defined in Listing~\ref{lst:MMT-PropLogSyn}
by a type constant for individuals (Listing~\ref{lst:LogSynDT}).
In this logic, we can define the domain theory (also Listing~\ref{lst:LogSynDT}),
which simply consists of constants such as \mmtinline{joan_DT} or \mmtinline{love_DT}.
As notation, we introduce \mmtinline{joan'} etc., following the convention in
NL semantics that the meaning of \str{Joan} is \mmtinline{joan'}.

\begin{center}
\begin{minipage}[t]{0.5\textwidth}
\begin{lstlisting}[language=MMT]
theory LogicSyntax : ur:?LF =
    include ?PropLogicSyntax ❙
    individual : type ❘ # ι ❙
❚
\end{lstlisting}
\end{minipage}\hfill
\begin{minipage}[t]{0.5\textwidth}
\begin{lstlisting}[language=MMT]
theory LifeDT : ?LogicSyntax =
    joan_DT : ι         ❘ # joan' ❙
    mary_DT : ι         ❘ # mary' ❙
    run_DT  : ι ⟶ o     ❘ # run' ❙
    love_DT : ι ⟶ ι ⟶ o ❘ # love' ❙
❚
\end{lstlisting}
\end{minipage}
    \captionof{lstlisting}{Logic and domain theory for the \gfinline{Life} grammar.}\label{lst:LogSynDT}
\end{center}

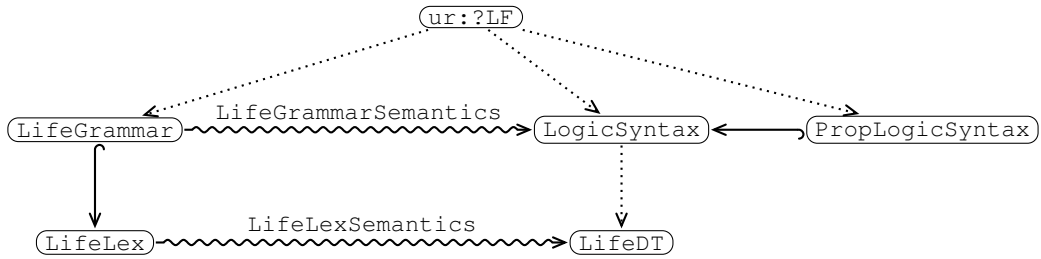
\begin{figure}\centering
    \begin{tikzpicture}[xscale=1,yscale=1]
    \node[thy] (lf) at (0, 2.5) {\theorynode{ur:?LF}};

    \node[thy] (lg) at (-5.0, 1.0) {\theorynode{LifeGrammar}};
    \node[thy] (ls) at (2.0, 1.0) {\theorynode{LogicSyntax}};
    \node[thy] (pls) at (6.0, 1.0) {\theorynode{PropLogicSyntax}};

    \node[thy] (ll) at (-5.0, -0.5) {\theorynode{LifeLex}};
    \node[thy] (ldt) at (2.0, -0.5) {\theorynode{LifeDT}};

    \draw[meta](lf) -- (lg);
    \draw[meta](lf) -- (ls);
    \draw[meta](lf) -- (pls);
    \draw[include](lg) -- (ll);
    \draw[meta](ls) -- (ldt);

    \draw[include](pls) -- (ls);
    
    \draw[view] (lg) -- node[above]{\theorynode{LifeGrammarSemantics}} (ls);
    \draw[view] (ll) -- node[above]{\theorynode{LifeLexSemantics}} (ldt);
    \end{tikzpicture}
    \caption{Semantics construction for the \gfinline{Life} example using \MMT's views.}\label{fig:LiveSemanticsConstruction}
  \end{figure}

Now that we have a target logic and a domain theory, we can define the semantics construction
as a modular view in \MMT.
Listing~\ref{lst:MMT-LifeSemantics-Split} represents it in two stages according to the grammar/lexicon partition of the source theory: 
The \mmtinline{LifeGrammarSemantics} view maps all constants from \mmtinline{LifeGrammar} to objects in \mmtinline{LogicSyntax}.
The type constants \mmtinline{Stmt}, \mmtinline{Person} and \mmtinline{Action} are mapped to propositions, individuals and unary predicates respectively.
\mmtinline{act} is mapped to a simple function application (with the argument order reversed) and \mmtinline{and} to the conjunction of \mmtinline{LogicSyntax} (recall that \mmtinline{[a,b]a∧b} stands for $\lambda a.\lambda b.a \land b$).
Since \mmtinline{LifeLex} includes \mmtinline{LifeGrammar}, all included constants must be mapped somewhere as well.
We can do this by simply including the \mmtinline{LifeGrammarSemantics} view described above.
In this small example, most new constants introduced in \mmtinline{LifeLex} have a corresponding element
in the domain theory. The only exception is \mmtinline{loveOneself}, which can be
described in terms of \mmtinline{love'}.
Figure~\ref{fig:LiveSemanticsConstruction} provides an overview of the theories
and views we just defined.
\MMT verifies the completeness of the semantics construction (i.e. that every element of the grammar
gets mapped to something)
by checking the totality of the views.

\begin{center}
\begin{minipage}[t]{.5\textwidth}
\begin{lstlisting}[language=MMT]
view LifeGrammarSemantics :
        ?LifeGrammar -> ?LogicSyntax =
    Stmt   = o ❙
    Person = ι ❙
    Action = ι ⟶ o ❙

    act = [pers, action] action pers ❙
    and = [a,b] a ∧ b ❙
❚
\end{lstlisting}
\end{minipage}\hfill
\begin{minipage}[t]{.5\textwidth}
\begin{lstlisting}[language=MMT]
view LifeLexSemantics :
        ?LifeLex -> ?LifeDT =
    include ?LifeGrammarSemantics ❙
    joan        = joan' ❙
    mary        = mary' ❙
    run         = run' ❙
    loveOneself = [x] love' x x ❙
    love        = love' ❙
❚
\end{lstlisting}
\end{minipage}
    \captionof{lstlisting}{Semantics construction for the \gfinline{Life} example using views into the domain theory.}\label{lst:MMT-LifeSemantics-Split}
\end{center}

Now we have everything we need to do the semantics construction for natural language utterances
of our small \gfinline{Life} fragment of the English language.
For example, we can parse the sentence \str{Joan loves herself} with \GF, which results
in the AST \gfinline{act joan loveOneself}.
Semantics construction just applies the view \mmtinline{LifeLexSemantics} from Listing~\ref{lst:MMT-LifeSemantics-Split} to the \ommt term \mmtinline{act joan loveOneself}.
\MMT computes \mmtinline{([pers,action]action pers) joan' [x]love' x x} and simplifies it to \mmtinline{love' joan' joan'}.

We end this section with a remark on truth conditions:
In \MMT the truth conditions of a statement can be represented as an \MMT theory.
For example, the truth conditions of \str{Mary runs and Joan runs} (or \mmtstyle{run' mary' $\land$ run' joan'}) could be represented as a \ommt theory with the axioms \mmtstyle{a1 : $\vdash$ run' mary'} and \mmtstyle{a2 : $\vdash$ run' joan'}.
Note that these axioms directly induce a Herbrand model, which is useful for many NLU systems.
It then remains to show that the statement indeed follows from the truth conditions, which
we can verify by giving a natural deduction proof -- i.e. a term in the theory
\mmtinline{PropLogicND} on the right of Listing~\ref{lst:MMT-PropLogSyn} -- which can be verified by the \MMT type-checker.

\section{Semantic Analysis}\label{sec:semAnal}
% While the semantics construction is compositional, 
% the semantic analysis often includes non-compositional inference steps.
% An easy approach to inference would be natural deduction.
% A natural deduction calculus can easily be described in \MMT via the Curry-Howard isomorphism.
% \GLF/\MMT would then serve as a proof-checker, but a human would have to construct
% the proofs.
% While this may be good enough from a philosophical point of view,
% it is insufficient for NLU systems which need automated inference.

At the end of the previous section we have described how truth conditions can be represented as theories containing some axioms that form a Herbrand model.
For an NLU system, these Herbrand models should of course be generated automatically.
This is possible with tableau calculi  which provide automated theorem proving and (Herbrand) model generation.

Following \cite{KohKol:ttmlu00} let us consider the following mini-discourse: \str{Mary is married to John. Her husband is not in town.}
After the semantic analysis it should be clear that John is not in town (indeed this is one of the truth conditions).
The result of the semantics construction could be the following expression in
sorted first-order logic:
\log{$married'(mary', john') \land \exists X_\mathbb{M}, Y_\mathbb{F}. (husband'(X, Y) \land \neg inTown'(X))$}
where $\mathbb{M}$ is a sort for male people and $\mathbb{F}$ a sort for female people.
These sorts are useful, because they correspond to English pronouns
making it clear that e.g. \str{her} cannot refer to \str{John}.
To infer that \str{her husband} is \str{John}, we need world knowledge
that describes married men as husbands and imposes monogamy:
\log{$\forall X_\mathbb{F}, Y_\mathbb{M}.married'(X, Y) \Rightarrow husband'(Y, X) \land (\forall Z_\mathbb{M}.
husband'(Z, X) \Rightarrow Z = Y)$}.
Model generation with a tableau calculus would then result in
the following facts: $married'(mary', john')$, $husband'(john', mary')$, $\neg inTown'(john')$
along with various negative facts in different branches (e.g. $\neg married'(mary', mary')$).

\subsection{Handling Natural Language Ambiguity}
Natural language is very prone to ambiguity.
One cause of ambiguity is \myemph{lexical ambiguity}:
one word can have several meanings (or, the other way around: different words can be spelt the same).
For example, in the sentence \str{Mary works at a bank}, the word \str{bank} could either refer
to a financial institution or to a geographical feature (river bank).
In \GF, this can be handled by providing two different entries in the abstract syntax --
\gfinline{bank_institute} and \gfinline{bank_river}, which are both linearized to the string \str{bank} in the concrete grammar for English. 
Parsing \str{Mary works at a bank} then results two different ASTs -- one with each meaning of the word \str{bank}.
Semantic analysis with world knowledge about legal working conditions might then indicate a preference for the ``financial institution reading''. 

\begin{figure}
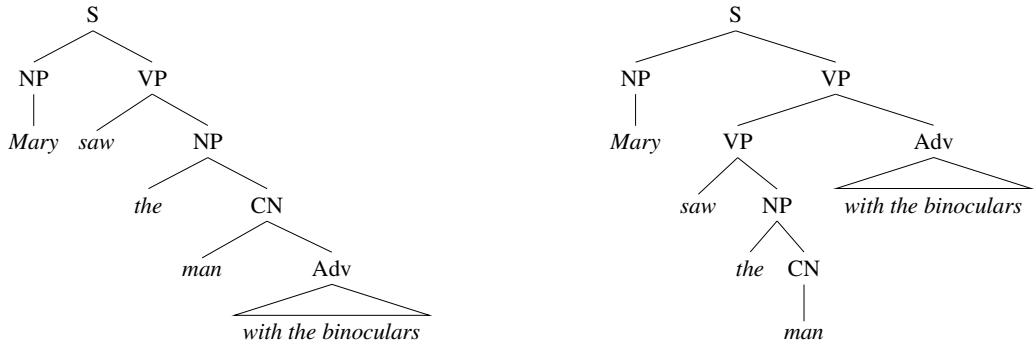

  \begin{minipage}[t]{0.5\textwidth}
    \scalebox{0.8}{
      \Tree [ .S [ .NP \textit{Mary} ] [ .VP \textit{saw} [ .NP \textit{the} [ .CN \textit{man} [ .Adv \edge[roof]; \textit{with the binoculars} ] ] ] ] ]
        }
      \end{minipage}\hfill
      \begin{minipage}[t]{0.5\textwidth}
        \scalebox{0.8}{
          \Tree [ .S [ .NP \textit{Mary} ] [ .VP [ .VP \textit{saw} [ .NP \textit{the} [ .CN \textit{man} ] ] ] [ .Adv \edge[roof]; \textit{with the binoculars} ] ] ]
        }
      \end{minipage}
    \caption{Two different (simplified) parse trees for the same sentence.}\label{fig:structAmbig}
\end{figure}

Another type of ambiguity is \myemph{structural ambiguity}, which means that the sentence structure is ambiguous.
For example, in the sentence \str{Mary saw the man with the binoculars},
the phrase \str{with the binoculars} could either refer to the verb \str{saw}
or to the object \str{the person}.
\GF handles this ambiguity also by creating one AST for each reading (Figure~\ref{fig:structAmbig}).

\GLF propagates ambiguity by applying semantics construction to each AST and passing the results on to semantic analysis.
The tableau machine from~\cite{KohKol:ttmlu00}, for example, would be given a disjunction of all terms, which creates one branch for each reading.
It then saturates the tableau inferentially with the available world knowledge until a resource criterion is met, and continues with all remaining open branches.
\cite{KohKol:ttmlu00} considers the open branches of a tableau, after the machine has been fed with a discourse sentence-wise, as the \textbf{semantic ambiguity}  of the discourse -- after all, they induce Herbrand models of the discourse.
Tableaux machines can be implemented easily on top of the \MMT data structures. 
We have experimented with a simplified tableau machine that can handle the
\mmtinline{LogicSyntax} described in Listing~\ref{lst:LogSynDT}.
An example output is shown at the end of Section~\ref{sec:GLF-appl}.

Finally, note that the syntactic ambiguity described above does not necessarily lead to
semantic ambiguity.
{
    \newcommand{\sem}[1]{\llbracket #1 \rrbracket}
    For example, the sentence \str{A and B and C} may be structurally ambiguous and result in the terms 
    \log{$(\sem{A} \land \sem{B}) \land \sem{C}$} and 
    \log{$\sem{A} \land (\sem{B} \land \sem{C})$}.
    However, the terms are equal given the associativity of \log{$\land$},
    i.e. there is no semantic ambiguity.
}

% \subsection{Simplification}
% After applying the semantics construction to a sentence like
% \str{Joan loves Mary or Mary loves Joan}, we are left with
% a term that still contains $\lambda$-expressions:
% 
% \mmtstyle{(([pers,action]action pers) joan (love mary))$\lor$(([pers,action]action pers) mary (love joan))}
% 
% \noindent What we actually expected was an expression in predicate logic (without quantifiers).
% By applying $\beta$-reduction, we can obtain an expression of the desired form:
% 
% \mmtstyle{((love mary) joan)$\lor$((love joan) mary)}
% 
% \noindent However, further simplification is still possible.
% Recall that we provided a definition of disjunction in terms of
% conjunction and negation (Listing~\ref{lst:MMT-PropLogSyn}).
% It is therefore possible to further simplify the expression with definition expansion:
% 
% \mmtstyle{$\neg$(($\neg$((love mary) joan))$\land$$\neg$((love joan) mary))}
% 
% While this may not be desirable for every application,
% it is for example useful for the tableau inference mentioned
% in Section~\ref{sec:semAnal} as it reduces the number of rules required.
% 
% \GLF supports any of the simplification levels described above.

\section{Examples}\label{sec:examples}
The running example based on the \gfinline{Life} fragment was very trivial and
does not really justify the efforts of combining \MMT and \GF.
Therefore, we will showcase two more fragments that have a more interesting target logic and semantics construction.
In Section~\ref{sec:fol} we will show a Montague-style variation of the \gfinline{Life} fragment which
contains quantifiers.
In Section~\ref{sec:modal}, we will use modal logic to model propositional attitudes.

\subsection{\label{sec:fol}Montague-Style Quantifiers}
\myemph{Noun phrases} are natural-language phrases that can serve as the subject or object of a verb.
In the \gfinline{Life} grammar, we represented them with the category \gfinline{Person}.
In this example, we will modify the grammar to support more complex noun phrases.
In particular, we will be able to talk about several people (\str{John and Mary})
and to quantify over the set of all people (\str{someone} or \str{everyone}).
This makes the semantics construction more challenging.
The new abstract syntax is shown in Listing~\ref{lst:GF-Quantifiers} on the left.
Note that we switched to the more syntactic (linguistic) category names \gfinline{NP} (noun phrase),
\gfinline{VP} (verb phrase), \gfinline{V2} (transitive verb) and \gfinline{S} (sentence).
We can use \GF's resource grammar library~\cite{RGL:on}
-- using an \gfinline{open} clause --
to create the English concrete syntax as
demonstrated in Listing~\ref{lst:GF-Quantifiers} on the right.
It provides all the language-specific rules we need, which means that we do not need to deal
with the word order and endings ourselves.
While it may not be very useful for our small example fragments,
it can simplify the implementation of larger fragments significantly.

\begin{figure}
    \begin{minipage}[t]{0.5\textwidth}
\begin{lstlisting}[language=GF]
abstract Quantified = {
  cat
    NP; VP; V2; S;
  fun
    applyObject  : V2 -> NP -> VP ;
    makeSentence : NP -> VP -> S ;
    and_NP       : NP -> NP -> NP ;
    everyone     : NP ;
    someone      : NP ;
}
\end{lstlisting}
    \end{minipage}\hfill
    \begin{minipage}[t]{0.5\textwidth}
\begin{lstlisting}[language=GF]
concrete QuantifiedEng of Quantified =
  open SyntaxEng,ParadigmsEng,DictEng in {
  lincat  -- map to RGL categories
    NP = NP; VP = VP; V2 = V2; S = S;
  lin
    applyObject v2 np = mkVP v2 np ;
    makeSentence n v = mkS (mkCl n v) ;
    and_NP a b = mkNP and_Conj a b ;
    everyone = everyone_NP ;
    someone = someone_NP ;
}
\end{lstlisting}
    \end{minipage}
    \captionof{lstlisting}{\label{lst:GF-Quantifiers}Variation of the \gfinline{Life} grammar that supports quantifiers (\str{everyone}, \str{someone}).}
\end{figure}

We will use first-order logic for the target representation.
The sentence \str{John loves everyone}
could then be represented as \log{$\forall x. love'(john', x)$}.
\mmtinline{FOL_Syntax} (Listing~\ref{lst:MMT-FOLSyn}) is an extension of the \mmtinline{LogicSyntax}
(defined in Listing~\ref{lst:LogSynDT}) with quantifiers using higher-order abstract syntax.
An extension for first-order natural deduction calculus is straightforward and well-understood. 

\begin{figure}[h]
\begin{lstlisting}[
    language=MMT,
    caption={First-order logic as an extension of \mmtinline{LogicSyntax}.},
    label={lst:MMT-FOLSyn}]
theory FOL_Syntax : ur:?LF =
    include ?LogicSyntax ❙
    forall : (ι ⟶ o) ⟶ o ❘ # ∀ 1 ❙
    exists : (ι ⟶ o) ⟶ o ❘ # ∃ 1 ❘ = [p] ¬ (∀ [x] (¬ p x)) ❙
❚
\end{lstlisting}
\end{figure}

The tricky part is now the semantics construction.
Consider the sentence \str{John and Mary run}, which should result in
the expression \log{$run'(john') \land run'(mary')$}.
Since the semantics construction has to be compositional, we
need to construct this expression from the meaning of \str{John and Mary} and the meaning of \str{run}.
The standard solution in a Montagovian framework~\cite{Montague:tptoqi73}
is \textbf{type raising}\footnote{
    The reader may be familiar with \myemph{continuation-passing style}, which is a similar concept in
    programming languages.
}:
by defining the meaning of \str{John and Mary} as \log{$\lambda p.p(john') \land p(mary')$},
we can compositionally define the meaning of \str{John and Mary run} as
\log{$(\lambda p.p(john') \land p(mary')) run'$}, which can be $\beta$-reduced to the desired expression
\log{$run'(john') \land run'(mary')$}.
This means that the type of noun phrases is now $(\iota \rightarrow o) \rightarrow o$.
Another way to look at this is that the meaning of a noun phrase is the set of its properties.
This approach results in mappings for the semantics construction like these:

\begin{lstlisting}[language=MMT]
    mary = [p : ι ⟶ o] p mary' ❙
    everyone = [p : ι ⟶ o] ∀ ([x : ι] p x) ❙
    and_NP = [np1, np2] [p : ι ⟶ o] (np1 p) ∧ (np2 p) ❙
\end{lstlisting}
% \mmtstyle{mary = [p : $\iota$ \jraa{} o] p mary' \jDD} \newline
% \mmtstyle{everyone = [p : $\iota$ \jraa{} o] $\forall$ ([x : $\iota$] p x) \jDD}\newline
% \mmtstyle{and_NP = [np1, np2] [p : $\iota$ \jraa{} o] (np1 p) $\land$ (np2 p) \jDD}

% Original:
% \noindent With some more type raising for verbs it is even possible to handle transitive verbs.
% Combining all this, we can obtain the meaning of sentences like \str{John and Mary love everyone}:
% 
% \lstinputlisting[language=MMT]{snippets/longquantifiedexample.mmt}
% 
% \noindent which can be $\beta$ reduced to
% 
% \mmtstyle{$\forall$[x:$\iota$](love' john' x)$\land$(love' mary' x)}

In order to handle transitive verbs like \str{love},
it is necessary to apply type raising to the verbs as well:
by providing verb phrases with noun phrases as arguments rather than individuals,
their type becomes $((\iota \rightarrow o) \rightarrow o) \rightarrow o$
where $(\iota \rightarrow o) \rightarrow o$ is the type of a noun phrase.
Similarly, transitive verbs get two noun phrases as arguments, resulting in the type
$((\iota \rightarrow o) \rightarrow o) \rightarrow ((\iota \rightarrow o) \rightarrow o) \rightarrow o$.
This allows us to define the semantics of verbs in the following way:

\begin{lstlisting}[language=MMT]
    run = [np] np run' ❙
    loveOneself = [np] np ([x : ι] (love' x) x) ❙
    love = [np1, np2] np1 ([x : ι] np2 ([y : ι] love' y x)) ❙
\end{lstlisting}

Note that \mmtstyle{$\text{run'}\;:\;\iota \rightarrow o$} and
\mmtstyle{$\text{love'}\;:\;\iota \rightarrow \iota \rightarrow o$} are constants from the domain theory (we can reuse the domain theory of the \gfinline{Life} example in Listing~\ref{lst:LogSynDT}).

It is now straightforward to define the semantics construction for
\gfinline{applyObject} and \gfinline{makeSentence}:

\begin{lstlisting}[language=MMT]
    applyObject = [v2, np] v2 np ❙
    makeSentence = [np, vp] vp np ❙
\end{lstlisting}

With all this, we can parse the example sentence \str{John and Mary love everyone}
to obtain the AST \gfinline{makeSentence (and_NP john mary) (applyObject love everyone)}.
Applying the semantics construction, we obtain the expression

\begin{lstlisting}[language=MMT]
([np,vp]vp np) (([np1,np2][p:ι⟶o](np1 p)∧(np2 p)) ([p:ι⟶o]p john') [p:ι⟶o]p mary')
(([v2,np]v2 np) ([np1,np2]np1 [x:ι]np2 [y:ι]love' y x) [p:ι⟶o]∀[x:ι]p x)
\end{lstlisting}

\noindent which $\beta$-reduces to 
\mmtstyle{$\forall$[x:$\iota$](love' john' x)$\land$(love' mary' x)}
as desired.

\subsection{\label{sec:modal}(Multi) Modal Logic}
In this example, we will use multi modal logic with the following modalities:
\begin{compactitem}
    \item \myemph{deontic} modality, expressing that something is obligatory or permitted
    \item \myemph{epistemic} modality, expressing that someone believes something to be true or possibly true
\end{compactitem}
\noindent For an introductory discussion of propositional attitudes and modal logics in natural language semantics we refer the reader to~\cite{Gamut:llmillg91}.
This text also discusses a plethora of other phenomena and logics, which we could have used as examples.
Indeed, all these logics -- and their combinations -- are a good validation
of the necessity of a tool like \GLF.

We can describe this logic in \MMT by extending the \mmtinline{LogicSyntax} from Listing~\ref{lst:LogSynDT}.
Listing~\ref{lst:MMT-ModLogSyn} shows the syntax for multi modal logic in general and for our
specific case with deontic and epistemic modalities.
As a notation we introduce \mmtstyle{$\llbracket$m$\rrbracket$} for the box operator with
modality \mmtstyle{m} and \mmtstyle{$\llangle$m$\rrangle$} for the diamond operator with modality
\mmtstyle{m}.

\begin{figure}[h]\centering
    \begin{minipage}[t]{0.5\textwidth}
\begin{lstlisting}[language=MMT]
theory MultiModalLogic : ur:?LF =
    include ?LogicSyntax ❙
    modality : type     ❘ # μ ❙
    box : μ ⟶ o ⟶ o     ❘ # ⟦ 1 ⟧ 2 ❙
    diamond : μ ⟶ o ⟶ o ❘ # ⟪ 1 ⟫ 2
        ❘ = [mod] [φ] ¬ ⟦ mod ⟧ ¬ φ ❙
❚
\end{lstlisting}
    \end{minipage}\hfill
    \begin{minipage}[t]{0.5\textwidth}
\begin{lstlisting}[language=MMT]
theory DEModalLogic : ur:?LF =
    include ?MultiModalLogic ❙
    deontic   : μ     ❘ # d ❙
    epistemic : ι ⟶ μ ❘ # e 1 ❙
❚
\end{lstlisting}
    \end{minipage}
    \captionof{lstlisting}{Multi modal logic syntax and an extension for deontic-epistemic modal logic.
        \label{lst:MMT-ModLogSyn}}
\end{figure}

This way, we can express the meaning of e.g. \str{John is not allowed to run} as \mmtstyle{$\neg$$\llangle$d$\rrangle$(run' john')} where \mmtstyle{d} is the deontic modality.
Similarly, we can express the meaning of \str{Mary believes that John is happy} with epistemic modality: \mmtstyle{$\llbracket$e mary'$\rrbracket$(happy' john')}.
% In \GF, this sentence is constructed with the rule
% 
% \gfinline{believeThat : Polarity -> Person -> S -> S ;}
% 
% \noindent where \gfinline{Polarity} is positive or negative to distinguish between \str{believes that} and \str{doesn't believe that}.
% In the semantics construction view, the meaning of \gfinline{believeThat} is defined as
% 
% \mmtstyle{believeThat = [pol,pers,s] pol ( $\llbracket$ e pers $\rrbracket$ s ) \jDD}
% 
% \noindent where \mmtstyle{e$\;\;$:$\;\;$$\iota$ \jraa{} $\mu$} constructs the epistemic modality for a person.
% An example of a sentence using both modalities would be
% 
% \str{John doesn't believe that Mary has to run}
% 
% \noindent which would result in the term
% 
% %\mmtstyle{$\neg$ $\llbracket$ e john' $\rrbracket$ $\llbracket$ d $\rrbracket$ run' mary'}
% \mmtstyle{$\neg$$\llbracket$(e john')$\rrbracket$$\llbracket$d$\rrbracket$(run' mary')}

In \GF, we can describe \str{be allowed to} and \str{have to} as verb phrase modifiers (\gfinline{VpModifier}),
while \str{Mary believes that} can be described as a sentence modifier (\gfinline{SModifier}).
To handle negations, we can introduce the category \gfinline{Pol} (for polarity),
indicating whether a sentence is negated.
Listing~\ref{lst:Modal} shows the abstract syntax in \GF.
For example, \str{John doesn't run} would be parsed into the AST \gfinline{makeS neg john run}
and \str{John has to run} would result in \gfinline{makeS pos john (modifyVP pos have_to run)}.

\begin{figure}[h]\centering
    \begin{minipage}[t]{0.5\textwidth}
\begin{lstlisting}[language=GF]
abstract Modal = {
  cat
    S ; VP ; Person;
    VpModifier ; SModifier;
    Pol; -- negative/positive polarity
  fun
    pos, neg : Pol;
    makeS : Pol->Person->VP->S;
    modifyVP : Pol->VpModifier->VP->VP;
    modifyS : Pol->SModifier->S->S;
    be_allowed_to : VpModifier;
    have_to : VpModifier;
    believe : Person->SModifier;
}
\end{lstlisting}
    \end{minipage}\hfill
    \begin{minipage}[t]{0.5\textwidth}
\begin{lstlisting}[language=MMT]
view ModalSemantics : ?Modal -> ?ModalDT =
  S = o ❙ VP = ι ⟶ o ❙ Person = ι ❙
  VpModifier = o ⟶ o ❙ SModifier = o ⟶ o ❙
  Pol = o ⟶ o ❙

  pos = [φ] φ ❙
  neg = [φ] ¬ φ ❙
  makeS = [pol,pers,vp] pol (vp pers) ❙
  modifyVP = [pol,m,vp] [x] m(pol(vp x)) ❙
  modifyS = [pol,m,s] pol (m s) ❙
  be_allowed_to = [φ] ⟪ d ⟫ φ ❙
  have_to = [φ] ⟦ d ⟧ φ ❙
  believe = [pers] [s] ⟦ e pers ⟧ s ❙
❚
\end{lstlisting}
    \end{minipage}
    \captionof{lstlisting}{\label{lst:Modal} The \GF abstract syntax along with the semantics construction.}
\end{figure}

The semantics construction is now rather straightforward (Listing~\ref{lst:Modal}):
sentences are propositions, verb phrases are unary predicates and persons are individuals.
The modifiers \gfinline{VpModifier} and \gfinline{SModifier} have the type \mmtstyle{$o \rightarrow o$},
which allows us to define, e.g., \gfinline{be_allowed_to} as \mmtstyle{$\lambda \varphi.\llangle\text{d}\rrangle \varphi$}, where \mmtstyle{$\llangle\text{d}\rrangle$} is the diamond operator with deontic modality.
The polarities can also be expressed as functions on propositions:
positive polarity is the identity function while negative polarity is negation.
These different components are combined with the functions \gfinline{makeS}, \gfinline{modifyVP}
and \gfinline{modifyS}, which simply apply the different components to each other.
For example, \gfinline{makeS} is mapped to \mmtinline{[pol,pers,vp] pol (vp pers)}.
With this, the semantics construction of the example sentence \str{John doesn't run}
would result in the term \mmtstyle{$\neg$(run' john')}.
An example sentence using modalities would be
\str{John doesn't believe that Mary has to run},
which results in the term
\mmtstyle{$\neg$$\llbracket$(e john')$\rrbracket$$\llbracket$d$\rrbracket$(run' mary')}.

\section{Conclusion and Future Work}\label{sec:concl}
We have presented \GLF, a simple framework for experimenting with natural-language semantics and developing Montagovian fragments.
It is -- to our knowledge -- the first framework that allows to implement the entire pipeline from language parsing to the semantic analysis in a declarative way.
The \GLF system reduces the creation of a symbolic NLU application to the following three steps:
\begin{compactenum}
\item Write a \GF grammar (abstract syntax + possibly multiple concrete syntaxes), possibly re-using large pieces of the \GF resource grammer. 
\item Define the target logic and domain theory, and define the semantics construction view in \MMT using the \GLF-generated language theory and a view stub. 
\item Implement some form of semantic analysis and application logic.
\end{compactenum}
We have tested this workflow with several examples, including the ones described in this paper.
The most complex example was the record $\lambda$-calculus for complex noun meanings from~\cite{PinKoh:flfdt00}.
We have also used \GLF as an educational tool in a course on logic-based natural language semantics at FAU Erlangen-N\"urnberg. 

To increase \GLF's usefulness, we are planning to work on the following points:
\begin{compactenum}
\item Providing a unified interface for the grammar and semantics development.
    Concretely, we started work on a Jupyter kernel for \GLF, extending our \GF Kernel (see~\cite{GFKernel}).
    Our goal is that the \GF grammar, the target logic, and the semantics construction view
    can all be implemented, tested, and documented in a single and coherent notebook.
\item Enabling \GLF to check that all $\lambda$ expressions introduced by type raising for the sake of compositionality get $\beta$-reduced away in semantics construction to ensure that the resulting expression is in the target logic.
\item Extending \MMT to calling an off-the-shelf theorem prover in the semantic analysis phase.
    This is particularly useful for pruning out readings in the style of~\cite{BlaBosKoh:iacs01} and thus reducing overall ambiguity. 
\item Looking into the reverse pipeline (logic-to-language translation),
    which could be a nice feature for displaying e.g. inferred results.
\item Adding support for regression testing and automated evaluation against a gold standard
    to facilitate realistic, corpus-driven development of fragments and ontologies.
\end{compactenum}
A small \GLF distribution along with the examples in this paper can be found at~\cite{GLFdemo}.

All in all, we hope that the \GLF system constitutes a tool that facilitates NLU development and experiments, and can thus kick-start a ``logic revival'' in  computational linguistics.
Like the systems it combines, it may also act as a bridge between the respective communities.

\nocite{Thomason:fp74}
\bibliography{specialbib}
\bibliographystyle{eptcs}
% \printbibliography
% \nocite{*}
% \bibliographystyle{eptcs}
% \bibliography{specialbib}

\end{document}